\documentclass[10pt,journal,compsoc]{IEEEtran}

\ifCLASSOPTIONcompsoc
  \usepackage[nocompress]{cite}
\else
  \usepackage{cite}
\fi

\ifCLASSINFOpdf
  \usepackage[pdftex]{graphicx}
  \graphicspath{{../pdf/}{../fig/}}
  \DeclareGraphicsExtensions{.pdf,.jpeg,.png}
\else
  \usepackage[dvips]{graphicx}
  \graphicspath{{../eps/}}
  \DeclareGraphicsExtensions{.eps}
\fi

\usepackage[ruled,vlined,linesnumbered]{algorithm2e}
\usepackage{amsmath, amssymb}

\usepackage[table,xcdraw]{xcolor}

\ifCLASSOPTIONcompsoc
  \usepackage[caption=false,font=footnotesize,labelfont=sf,textfont=sf]{subfig}
\else
  \usepackage[caption=false,font=footnotesize]{subfig}
\fi

\usepackage{url}
\usepackage{booktabs}
\usepackage{multirow}
\usepackage{boldface_pei}
\usepackage{makecell}

\usepackage{threeparttable}

\begin{document}

\title{A Greedy Strategy for Graph Cut}

\author{Feiping~Nie, Shenfei~Pei, Zengwei~Zheng, Rong~Wang, and~Xuelong~Li,~\IEEEmembership{Fellow,~IEEE}
\IEEEcompsocitemizethanks{
\IEEEcompsocthanksitem Corresponding author: Feiping Nie.
\IEEEcompsocthanksitem F.~Nie and S.~Pei are with the School of the Computer Science, and the School of Artificial Intelligence, OPtics and ElectroNics (iOPEN), 
Northwestern Polytechnical University, Xi’an 710072, Shaanxi, P. R. China.
E-mail: feipingnie@gmail.com, shenfeipei@gmail.com.
\IEEEcompsocthanksitem Z.~Zheng is with the School of Computer and Computing Science, Hangzhou City University, Hangzhou 310015, Zhejiang, P. R. China. E-mail: zhengzw@hzcu.edu.cn.
\IEEEcompsocthanksitem R.~Wang and X.~Li are with the School of Artificial Intelligence, OPtics and ElectroNics (iOPEN), the Key Laboratory of Intelligent Interaction and Applications (Ministry of Industry and Information Technology), and the School of Computer Science, Northwestern Polytechnical University, Xi’an 710072, Shaanxi, P. R. China. \protect\\
E-mail: wangrong07@tsinghua.org.cn, li@nwpu.edu.cn.
}}

\markboth{Journal of \LaTeX\ Class Files,~Vol.~14, No.~8, August~2015}%
{Shell \MakeLowercase{\textit{et al.}}: Bare Demo of IEEEtran.cls for Computer Society Journals}
%



\IEEEtitleabstractindextext{%
\begin{abstract}
We propose a Greedy strategy to solve the problem of Graph Cut, called GGC. 
It starts from the state where each data sample is regarded as a cluster and 
dynamically merges the two clusters which reduces the value of the global objective function the most 
until the required number of clusters is obtained, 
and the monotonicity of the sequence of objective function values is proved.
To reduce the computational complexity of GGC, 
only mergers between clusters and their neighbors are considered. 
Therefore, GGC has a nearly linear computational complexity with respect to the number of samples.
Also, unlike other algorithms, due to the greedy strategy, the solution of the proposed algorithm is unique. 
In other words, its performance is not affected by randomness.
We apply the proposed method to solve the problem of normalized cut which is a widely concerned graph cut problem. 
Extensive experiments show that better solutions can often be achieved 
compared to the traditional two-stage optimization algorithm (eigendecomposition + $k$-means), 
on the normalized cut problem. 
In addition, the performance of GGC also has advantages compared to several state-of-the-art clustering algorithms.
\end{abstract}

}

\maketitle

\IEEEdisplaynontitleabstractindextext

%
\IEEEpeerreviewmaketitle

\IEEEraisesectionheading{\section{Introduction}\label{sec:introduction}}

%
%
%
%
\IEEEPARstart{C}{lustering}, as a fundamental task in machine learning, 
has gained increasing attention in recent years \cite{cc}. 
Many clustering algorithms have been proposed and applied to various fields, 
such as text analysis \cite{cluster_text}, face recognition \cite{cluster_face}, 
image segmentation \cite{cluster_seg}, etc.
Among these algorithms, spectral clustering, 
a family of algorithms based on eigenvalue decomposition, 
has garnered much research attention due to its tendency to produce superior experimental performance.

Generally, spectral clustering can be derived as an approximation to a graph partitioning problem. 
For instance, the method proposed in \cite{ncut2}, its corresponding graph cut problem is called normalized cut.
Let $\X = [\x_1, \cdots, \x_n] \in \mathbb{R}^{d \times n}$ be the dataset, $\G = <\V, \W>$ be an undirected weighted graph, where $\V$ represents the vertex set, and $\W \in \mathbb{R}^{n \times n}$ represents the weighted adjacency matrix.
$w_{ij} > 0$ if $\x_i \in \mathcal{N}(\x_j)$ or $\x_j \in \mathcal{N}(\x_i)$, $w_{ij} = 0$ otherwise. $\mathcal{N}(\x_i)$ is the set consisting of the $k$-nearest neighbors of $\x_i$.
The objective function of normalized cut can be expressed as follows:
\begin{equation}\label{ncut}
\min_{\Y \in \Phi^{n \times c}} Tr \left(  \left( \Y^T \D\Y \right) ^{-1} \Y^T \L\Y \right),
\end{equation}
where $\Phi^{n \times c}$ denotes the set of indicator matrices, $n$ and $c$ denote the number of samples and clusters, 
$\L$ is the unnormalized graph Laplacian matrix, $\L = \D-\W$, and $\D$ is the diagonal degree matrix.
The meaning it expresses is that the partition to be solved should make the similarity between clusters smaller. For more explanations, please refer to \cite{cluster_tur}.
Since its search space consists of discrete points, the problem of Eq.~\eqref{ncut} is difficult to solve. Therefore, in spectral clustering, the problem is solved in two stages.
\begin{itemize}
\item Embedding: Let $\F=\D^{1/2} \Y (\Y^T \D \Y )^{-1/2}$, and solve $\F$ by eigenvalue decomposition in a relaxed continuous domain, where only the orthogonal constraint $\F^T \F=\I$ is preserved.
\item Discretization: It is non-trival to obtain the division from the embedding. So we need to get the final clustering result with the help of other discretization methods, such as $k$-means \cite{kmeans} or spectral rotation \cite{SR}.
\end{itemize}
Other spectral clustering methods are mostly two-stage, such as Ratio-cut \cite{rcut}, Improved Spectral Rotation \cite{ISR}, multiway p-spectral clustering \cite{mpsc}, and so on.
From the above discussion, we know that: 
1. The time complexity is high because it involves the eigenvalue decomposition of the Laplacian matrix of size $n$ by $n$. 2. The solution obtained by the algorithm cannot be guaranteed to be the local optimal solution to problem~\eqref{ncut}. 
Although some methods \cite{joint1, joint2} integrate embedding and discretization together, the solution is still not guaranteed to be a local optimal.

\begin{figure*}[]
	\centering
	\includegraphics[width=0.7\textwidth]{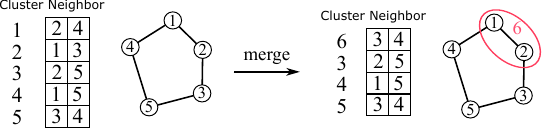}
  \caption{Diagram of the merge operation. Cluster 6 represents a new cluster obtained by merging 1 and 2. The neighbors of most clusters are unchanged, so a lot of information can be reused, which is very helpful for speeding up our algorithm.}\label{fig:demo}
	\label{fig:ny}
\end{figure*}

We propose a optimization method for the problem~\eqref{ncut}, based on the following observation: 
Given a graph $\G$ and a solution $\Y \in \Phi^{n \times (c+1) }$, there is always a solution $\hat{\Y} \in \Phi^{n \times c}$ 
obtained by merging two clusters in $\Y$ such that $f(\hat{\Y}) \leq f(\Y)$ 
where $f$ represents the objective function of the normalized cut.
We start from the state where each sample is a separate cluster and 
repeatedly merge the two clusters that decrease the global objective function the most, 
until the required number of clusters is reached as shown in Fig. \ref{fig:demo}.
We proposed a fast optimization algorithm for finding the two clusters to merge, 
with nearly linear time complexity w.r.t the number of samples. 

\IEEEpubidadjcol

This paper makes the following key contributions:
\begin{itemize}
\item We propose a new optimization algorithm for the problem of normalized cut 
that iteratively reduces the objective function value through greedy merging of cluster pairs, 
effectively instantiating a greedy approach
which can be seen as an instance of the greedy algorithm.
\item Neither eigenvalue decomposition nor singular value decomposition are involved in the proposed algorithm. 
\item GGC demonstrates linear computational complexity with respect to the number of samples, 
which means that it can be easily scalable to large-scale problems.
\item Unlike other algorithms, GGC initializes with the state where each point is treated as a cluster, 
and the solution obtained from it is unique obviating the need for multiple repeated runs.
\end{itemize}

\textbf{Notations:}
Throughout this paper, we denote
matrices by boldface uppercase letters and vectors by boldface lowercase letters.
For a matrix with size $d$ by $n$, $\M \in \mathbb{R}^{d \times n}$, 
we denote the $i$-th column and the $(i,j)$-th entry of it as $\m_i$ and $m_{ij}$ respectively, 
where $\m_i \in \mathbb{R}^{d}$ is a d-dimensional column vector.
$Tr(\M)$ represents the trace of matrix $\M$. 
We refer to a matrix as \textit{an indicator matrix} if it has exactly one element of $1$ in each row, 
with the remaining elements as $0$. 
$\Phi^{n \times c}$ denotes the set of indicator matrices with size $n$ by $c$.
$\mathbf{1}$ denotes a column vector of all $1$s, whose size depends on the context.

\section{Related Work}\label{sec:work}
Spectral clustering refers to a family of algorithms based on eigenvalue decomposition. 
Among many spectral clustering algorithms, 
the methods proposed by \cite{rcut, ncut, ncut2} are the most popular and receive more and more attention. 
However, most of these algorithms adopt a two-stage strategy and have a high computational complexity, 
as described in Section \ref{sec:introduction}.
Much work has focused on addressing these limitations.

The anchoring approach proposed in \cite{anchor1, anchor2} is a simple and effective way to speed up spectral clustering. 
These methods first select some sample points called anchors, then measure similarity between samples and anchors, 
and finally represent inter-sample weight matrix using the similarity.
In particular, the weight matrix, $\A \in \mathbb{R}^{n \times n}$, between samples can be expressed as
\begin{equation}
\A = \Z \mathbf{\Delta}^{-1} \Z^T,
\end{equation}
where $\Z \in \mathbb{R}^{n \times m}$ represents the similarity matrix between samples and anchors, 
$\mathbf{\Delta}$ is a diagonal matrix with $\Delta_{jj}=\sum_{i=1}^n z_{ij}$. 
Consequently, the eigenvectors of matrix $\A$ are then equivalent to the left singular vectors of matrix $\B=\Z \mathbf{\Delta}^{-1/2}$.
Based on the similarity matrix between samples and anchors, \cite{fcdmf} proposes an efficient optimization method 
without solving singular vectors where the discrete indicator matrix can be obtained directly, 
thus averting the need for subsequent discretization.

The Nystrom approximation \cite{Ny1} has also been widely adopted for accelerating spectral clustering. 
Vladymyrov et al \cite{Nyfast1} showed that an approximation that can be optimized efficiently 
can be obtained by cooperating the Nystrom formula as a constraint to the spectral clustering. 
This method can achieve lower approximation error with fewer landmarks.
Jia et al \cite{Nyfast2} showed the standard Nystrom eigenvectors are suboptimal. 
Consequently, they designed a new matrix factorization strategy for Nystrom spectral clustering, 
and proposed an efficient algorithm based on incomplete Cholesky decomposition to solve the eigenvectors.

To improve the solution quality of spectral clustering, many algorithms has been proposed.
According to the above discussion, the problem corresponding to solving eigenvectors in traditional spectral clustering is:
\begin{equation}
\min_{\F^T \F = \I} Tr\left( \F^T \D^{-1/2} \L \D^{-1/2} \F \right).
\end{equation}
Yu et al \cite{SR} showed that if $\F^*$ is optimal for the above problem, 
then so is $\F^* \R$, where $\R^T\R=\I$ is a rotation matrix. 
Therefore, discretizing $\F$ may be not the best choice. 
To this end, they learn $\R$ and $\Y$ simultaneously by
\begin{equation}\label{yyr}
\min_{\Y \in \Phi^{n \times c}, \R \in \mathbb{R}^{c \times c}, \R^T\R = \I}   \Vert \Y - \Y^* \R \Vert_F^2, 
\end{equation}
where $\Y^* = diag\left(\F^* (\F^*)^T \right)^{-1/2} \F^*$.
Since $\Y$ is an approximate solution, the optimum of Eq.~\eqref{yyr} may deviate from the ground truth. To this end, Chen et al \cite{ISR} proposed Improved Spectral Rotation (ISR). From the description of Section \ref{sec:introduction}, we know that $\F$ is an approximation of $\D^{1/2} \Y \left(\Y^T \D \Y \right)^{-1/2}$, so they learn $\Y$ and $\R$ simultaneously according to
\begin{equation}
\min_{\Y \in \Phi^{n \times c}, \R^T \R = \I}
\Vert  \D^{1/2} \Y \left( \Y^T \D \Y \right)^{-1/2} - \F^* \R \Vert_F^2.
\end{equation}
However, the above methods are two-stage. 
To this end, Pang et al \cite{joint1} proposed a joint model to simultaneously learn the embedding and the indicator matrix, which can be seen as an improved version of \cite{ISR}.
Likewise, Wang et al \cite{joint2} also proposed a framework where spectral embedding and 
discretization are learned simultaneously and proved that the rotation matrix $\R$ is unnecessary.
They also learned a low-dimensional representation matrix by label propagation, with help of this matrix,  the computational complexity is reduced significantly.

Because the merging operation of the proposed algorithm in this work is very common in Agglomerative Hierarchical Clustering (AHC), 
so a brief introduction to AHC is given below.
Unlike partition-based clustering algorithms, AHC adopts a bottom-up framework, 
it does not optimize a global objective function, but instead recursively merges clusters until all samples belong to one cluster. 
The key of AHC is designing an appropriate criterion to identify the closest clusters to merge.
Tabatabaei et al proposed an efficient hierarchical clustering algorithm, GANC (Greedy Agglomerative Normalized Cut \cite{GANC}).
GANC consists of the following three parts:
1. An agglomerative hierarchical clustering procedure where a criterion based on normalized cut is used. 
It aims to minimize the normalized cut loss at each layer.
2. Selecting the appropriate number of clusters based on the loss curve or prior knowledge.
3. Refinement. The hierarchical clustering procedure often gets trapped in local optima due to the greedy strategy. 
To address this, a refinement step is adopted where boundary samples are moved across clusters 
to further improve the clustering quality.
For more AHC algorithms, please refer to \cite{hc1, hc3, hc4, hc5}.

\section{The Proposed Model}
Our algorithm starts from the initial state where each sample is regarded as a cluster and reduces the value of the objective function of normalized cut by merging clusters. Because the number of clusters is reduced by $1$ after each merger, our algorithm always requires $t = n-c$ mergers, where $n$ and $c$ represent the number of samples and clusters, respectively.
To begin with, we rewrite the problem of the normalized cut as follows:
\begin{align}
&\min_{\Y \in \Phi^{n \times c}} 
Tr \left( \left( \Y^T \D \Y \right)^{-1} \Y^T \L \Y  \right), \\
\Leftrightarrow
&\min_{\Y \in \Phi^{n \times c}}  \sum_{k=1}^c \frac{\y_k^T \L \y_k}{\y_k^T \D \y_k}.
\end{align}
In our algorithm, a total of $t$ solutions are involved, for convenience, we denote them as
\begin{equation}
\Y^{(n)} \Rightarrow \Y^{(n-1)} \Rightarrow \cdots \Rightarrow \Y^{(c)},
\end{equation}
where 
$\Y^{(m)} \Rightarrow \Y^{(m-1)}$ means  $\Y^{m-1}$ obtained by merging two clusters in $\Y^{(m)}$, and
$\Y^{(m)}$ containing $m$ clusters represents the solution obtained after $n-m$ times of merging. 
In other words,
$\Y^{(n)}$ represents the initial solution, $\Y^{(c)}$ represents the final solution.

As discussed earlier, our algorithm is always looking for two clusters that merging them can decrease the objective function the most.
Defining
\begin{align}
l^{(m)}_{ij} &= \left( \y_i^{(m)} \right) ^T \L \y^{(m)}_j, i, j = 1, \cdots, m, \\
d^{(m)}_k    &= \left( \y_k^{(m)} \right) ^T \D \y^{(m)}_k, k = 1, \cdots, m.
\end{align}
Without loss of generality, we assume that  $\Y^{(m-1)}$ is obtained by combining two clusters $i$ and $j$. Then we have
\begin{align}
\delta^{(m)}_{ij} &= f\left( \Y^{(m)} \right) - f\left(  \Y^{(m-1)}  \right)  \label{YmYm1} \\
             &= \frac{ l_{ii}^{(m)}    }{  d_i^{(m)}   } 
              + \frac{ l_{jj}^{(m)}    }{  d_j^{(m)}   }
              - \frac{ l_{ee}^{(m-1)}  }{  d_e^{(m-1)} },  \label{fe}
\end{align}
where $f$ is the objective function of normalized cut, and $e$ represents the cluster formed by the combination of $i$ and $j$.
Since $\y_e^{(m-1)} = \y_i^{(m)} + \y_j^{(m)}$ and 
$\L$ and $\D$ are symmetric matrices, we have
\begin{align}
l^{(m-1)}_{ee} &= l^{(m)}_{ii} +  l^{(m)}_{jj} + 2l^{(m)}_{ij},  \\
d^{(m-1)}_e    &= d^{(m)}_i    +  d^{(m)}_j + 2 \left( \y^{(m)}_i \right) ^T \D \y^{(m)}_j.
\end{align}
Since $\D$ is a diagonal matrix and $\y_i$ and $\y_j$ are indicator vectors, we have
\begin{align}
\left( \y^{(m)}_i \right) ^T \D \y^{(m)}_j = \sum_{k=1}^n d_{kk} y^{(m)}_{ik} y^{(m)}_{jk}  = 0.
\end{align}
So far, Eq.~\eqref{fe} can be equivalently written as
\begin{equation}\label{yaLyb}
\begin{split}
\delta^{(m)}_{ij} =  \frac{ l_{ii}^{(m)}  }{  d_i^{(m)}   } +  \frac{ l_{jj}^{(m)}  }{  d_j^{(m)}   } - 
                      \frac{ l_{ii}^{(m)}  +  l_{jj}^{(m)} + 2l_{ij}^{(m)} }
                            { d_i^{(m)}   + d_j^{(m)}   }.
\end{split}
\end{equation}

Since 
\begin{align}
& d^{(m)}_i > 0, l^{(m)}_{ii} \geq 0, l^{(m)}_{ij} \leq 0,  \\
& \forall m = c, \cdots, n, i,j = 1, \cdots, m, i \not = j, 
\end{align}
the value of Eq. \eqref{yaLyb} is always greater than or equal to 0, and its value is equal to 0 if and only if
$l^{(m)}_{ii} = 0$, $ l^{(m)}_{jj} = 0$, and $l^{(m)}_{ij} = 0$.
Therefore, we have
\begin{equation}
f\left( \Y^{(n)} \right) \geq f\left( \Y^{(n-1)} \right) \geq, \cdots,  \geq f\left( \Y^{(c)} \right) \geq 0.
\end{equation}
In other words, the objective function values form a monotonic decreasing sequence bounded below by 0.

As discussed earlier, our algorithm is always looking for two clusters that merging them can decrease the objective function the most.
Therefore, our algorithm can be summarized in the following steps:
\begin{itemize}
\item \textbf{Step 0} Initialize $\Y^{(n)}$ where each cluster contains only one sample, and $m \leftarrow n$.
\item \textbf{Step 1} Compute the merge gains $\delta^{(m)}_{ij}$ between all cluster pairs. 
\item \textbf{Step 2} Find the maximum $\delta^{(m)}_{i^*j^*}$ and merge clusters $i^*$ and $j^*$ into one cluster. $m \leftarrow m - 1$
\item \textbf{Step 3} Repeat steps 1 and 2 until $m$ equals $c$.
\end{itemize}

We define:
\begin{align}
\hat{\mathbb{F}}^{(m)} &=  \{ \delta^{(m)}_{ij} \mid i,j=1, \cdots, m  \},
\end{align}
It can be seen that the core problem is to find the maximum of $\hat{\mathbb{F}}^{(m)}$, for $m=n, ..., c+1$.
Obviously, $ \vert \hat{\mathbb{F}}^{(m)} \vert = m^2$, so it takes $O(m^2)$ time to find the minimum.
That is, even excluding the time required to compute $\hat{\mathbb{F}}^{(m)}$, the overall computational complexity of our algorithm is 
$\sum_{m=c+1}^n O(m^2) = O(n^3)$,
which is very time-consuming.

\subsection{Make It Faster}
To reduce GGC's time complexity, two strategies are adopted: 
Neighbor Cluster Searching and a red-black tree. 
These strategies result in GGC having a near linear time complexity with respect to the number of samples.

\textbf{Neighbor Cluster Searching:}
If the size of the set $\hat{\mathbb{F}}^{(m)}$ can be reduced, there will be a gain in running time.
To this end, we first introduce the concept of the neighbor cluster.
For two clusters, the $i$-th cluster and the $j$-th cluster, we call them neighbors to each other, if
\begin{equation}
l^{(m)}_{ij} < 0.
\end{equation}
From the above formula, we can see that if the $i$-th cluster is not a neighbor of the $j$-th cluster, 
then for any sample in the $j$-th cluster, its neighbors are all not in the $i$-th cluster. 
Therefore, it does not make sense to merge the two clusters.
In GGC, only the first $k$ neighbors will be concerned, that is
\begin{equation}
\mathbb{F}^{(m)} = \{ \delta^{(m)}_{ij} \mid i=1, \cdots, m, j \in \mathcal{C}(i)  \},
\end{equation}
where $\mathcal{C}(i)$ denotes the first $k$ neighbors of the $i$-th cluster.
Let $k_1$ denote the average number of neighbors per cluster, then we have $\vert \mathbb{F}^{(m)} \vert = m k_1 \leq n k$.

There is significant overlap between the sets 
$\mathbb{F}^{(m)}$ and $\mathbb{F}^{(m-1)}$. 
To be more precise, the elements in subset 
$\{ \delta^{(m-1)}_{ij} \mid i \not= e \text{ and } j\not = e  \}$ where $e$ represents the newly generated cluster, have all appeared in
$\mathbb{F}^{(m)}$.
This means that we don't need to calculate each element of $\mathbb{F}^{(m)}$.

Suppose $\delta^{(m)}_{i^* j^*}$ is the smallest element in the set $\mathbb{F}^{(m)}$,
and $e$ is the cluster obtained by merging $i^*$ and $j^*$, we have 
\begin{align}
\mathcal{A}^{(m)}  &= \{ \delta^{(m)}_{i^*k} \mid k \in \mathcal{C}(i^*)  \} \cup \{ \delta^{(m)}_{j^*k} \mid k \in \mathcal{C}(j^*)  \}, \\
\mathbb{F}^{(m-1)} &= \left( \mathbb{F}^{(m)} \setminus \mathcal{A}^{(m)} \right) \cup \{ \delta^{(m-1)}_{ek} \mid k \in \mathcal{C}(e) \}. \label{ff1}
\end{align}

\textbf{Red-black Tree:}
The time-consuming steps of GGC are: 
a) Find the smallest element of $\mathbb{F}^{(m)}$. 
b) Generate $\mathbb{F}^{(m-1)}$ based on $\mathbb{F}^{(m)}$.
According to equation \eqref{ff1}, we know that $\mathbb{F}^{(m-1)}$ can be generated 
by performing insertion and deletion operations on  $\mathbb{F}^{(m)}$.
Therefore, we use a red-black tree to store the elements in  $\mathbb{F}^{(m)}$, 
which allows us to complete search, insertion, and deletion operations in $O(\log(n_e))$ time, 
where $n_e$ is the number of entries in the tree.
In our algorithm, the red-black tree has $n k_1$ elements after initialization,
where $k_1$ is the average of the number of neighbors of each cluster, 
After that, the number of elements in the tree becomes smaller and smaller.

The proposed algorithm to solve the problem of normalized cut is summarized in Algorithm 1.
Next, we make some comments on some important operations in Algorithm 1.
\begin{itemize}
\item The insertion (line 8 and 25), deletion (line 14 and 16), 
and search (line 10) involved in the algorithm are all built-in operations of the red-black tree, 
and their computational complexity is at most $O(log(n k_1))$.
\item It only takes $O(1)$ time to calculate $\delta^{(m)}_{ij}$ (line 7 and line 23), 
because the required variables $l^{(m)}_{ii}, l^{(m)}_{jj}, l^{(m)}_{ij}, d^{(m)}_i$ and $d^{(m)}_j$ are all known.
\end{itemize}

\begin{algorithm}[!t]
\KwData{$\L \in \mathbb{R}^{n \times n}$, $\D \in \mathbb{R}^{n \times n}$, 
the number of cluster, $c$.}
\KwResult{Clustering result, $\Y^{(c)}$.}
Initialize $\Y^{(n)}$, $\y^{(n)}_k = list(k), k=1, \cdots, n$\;
Initialize $l^{(n)}_{ij} = \L_{ij}, i=1,\cdots, n, j \in \mathcal{C}(i)$ \;
Initialize $d^{(n)}_{k} = \D_{kk}, k = 1, \cdots, n$\;
Initialize the red-black tree $\mathbb{F}^{(n)} = \varnothing$\; 
\For{$i = 1, \cdots, n$}{
  \For{$j \in \mathcal{C}(i)$}{
    Compute $\delta^{(n)}_{ij}$ by Eq.~\eqref{yaLyb}\;
    Add $(\delta^{(n)}_{ij}, i, j)$ to $\mathbb{F}^{(n)}$\;
	}
}
\For{$m=n, \cdots, c+1$}{
  Find the smallest element of $\mathbb{F}^{(m)}$, $\delta^{(m)}_{i^*j^*}$\;
  $\Y^{(m-1)} = \Y^{(m)}$\;
  $e = 2n + 1 - m$\;
  $\y^{(m-1)}_e = \y^{(m)}_{i^*} \cup \y^{(m)}_{j^*}$\;
  \For{$k \in \mathcal{C}(i^*)$}{
    Remove $(\delta^{(m)}_{i^*k}, i^*, k)$ from $\mathbb{F}^{(m)}$\;
  }
  \For{$k \in \mathcal{C}(j^*)$}{
    Remove $(\delta^{(m)}_{j^*k}, j^*, k)$ from $\mathbb{F}^{(m)}$\;
  }
  \For{$p \in \mathcal{C}(i^*) \cup \mathcal{C}(j^*)$}{
      $\mathcal{C}(p).remove(i^*, j^*)$\;
      $\mathcal{C}(p).insert(e)$\;
  }
  $\mathcal{C}(e) = \mathcal{C}(i^*) \cup \mathcal{C}(j^*)$\;
  $l^{(m-1)}_{ep} = l^{(m)}_{i^*p} + l^{(m)}_{j^*p}, p \in \mathcal{C}(e)$\;
  $d^{(m-1)}_e = d^{(m)}_{i^*} + d^{(m)}_{j^*}$\;
  $\mathbb{F}^{(m-1)} = \mathbb{F}^{(m)}$\;
  \For{$p \in \mathcal{C}(e) $}{
      Compute $\delta^{(m-1)}_{ep}$ by Eq.~\eqref{yaLyb}\;
      Add $(\delta^{(m-1)}_{ep}, e, p)$ to $\mathbb{F}^{(m-1)}$\;
  }
}
\caption{Accelerated algorithm for solving problem of normalized cut}
\label{algo1}
\end{algorithm}

\textbf{Computational complexity:}
Let $k_1$ be the average of the number of neighbors of each cluster, 
we need at most $O(nk_1\log(nk_1))$ time to initialize variables such as $\mathbb{F}^{(n)}$ and $\Y^{(n)}$.
The operation of finding the smallest element of $\mathbb{F}^{(m)}$ takes at most $O(\log(nk_1))$ time (line 10).
Since the set of indices of each cluster is implemented as a list, 
the merge between two clusters only involves the modification of pointers. So it takes $O(1)$ time (line 12).
For each $m$, there will be $k_1$ elements added to $\mathbb{F}^{(m)}$.
So, the insert operation takes at most $O(k_1\log(nk_1))$ time (line 22-24)
Therefore, the computational complexity of the proposed algorithm is $O((n-c)k_1 \log(nk_1))$.
Since $\log(n) \leq \log(nk_1) \leq 2\log(n)$, 
the computational complexity of GGC can be written as $O((n-c)k_1 \log(n))$.

\textbf{Space Complexity:} 
The memory is mostly occupied by variables $\L$, $\D$ and $\mathbb{F}^{(m)}$. 
Because $\L$ is a sparse matrix, it requires $O(nk_2)$ memory, where $k_2$ is the average number of non-zero elements in each row of $\L$.
Since $\D$ is a diagonal matrix, it requires $O(n)$ memory only. 
Since set $\mathbb{F}$ has at most $O(nk_1)$ elements, it requires $O(nk_1)$ memory.
Thus, the space complexity of GGC is $O(n(k_1+k_2))$.

%
%
%
%
%


\begin{table*}[]
\renewcommand{\arraystretch}{1.2}
\caption{The value of objective function on 16 real-world datasets. The best results are shown in bold.}
\label{tab:obj}
\centering
  \begin{tabular}{cccccccc}
  \toprule
  Datasets & $M_{1}$                    & $M_{2}$                    & $M_{3}$                     & $M_{4}$                    & $M_{5}$                    & $M_{6}$                     & $M_{7}$                    \\ \midrule
  TSC      & 11.744($\pm$0.622)         & 1.538($\pm$0.127)          & 0.691($\pm$0.137)           & 4.149($\pm$0.327)          & \textbf{0.916($\pm$0.016)} & 0.885($\pm$0.383)           & \textbf{0.537($\pm$0.007)} \\
  GGC      & \textbf{0.32}              & \textbf{1.459}             & \textbf{0.002}              & \textbf{2.181}             & 0.951                      & \textbf{0.781}              & 0.622                      \\ \midrule
  Datasets & $M_{8}$                    & $M_{9}$                    & $M_{10}$                    & $M_{11}$                   & $M_{12}$                   & $M_{13}$                    & $M_{14}$                   \\ \midrule
  TSC      & \textbf{0.177($\pm$0.271)} & \textbf{0.365($\pm$0.005)} & 1.925($\pm$0.279)           & 12.195($\pm$0.503)         & 2.091($\pm$0.080)          & 38.118($\pm$1.658)          & \textbf{2.422($\pm$0.200)} \\
  GGC      & 0.335                      & 0.407                      & \textbf{0.11}               & \textbf{0.295}             & \textbf{1.831}             & \textbf{5.837}              & 2.762                      \\ \midrule
  Datasets & $M_{15}$                   & $M_{16}$                   & $L_{1}$                     & $L_{2}$                    & $L_{3}$                    & $L_{4}$                     &                            \\ \midrule
  TSC      & 0.759($\pm$0.104)          & \textbf{0.827($\pm$0.011)} & 34.023($\pm$1.163)          & 0.546($\pm$0.060)          & \textbf{3.923($\pm$0.113)} & \textbf{9.518($\pm$0.169)}  &                            \\
  GGC      & \textbf{0.625}             & 0.925                      & \textbf{25.715}             & \textbf{0.477}             & 4.293                      & 10.896                      &                            \\ \bottomrule
  \end{tabular}
\end{table*}

\subsection{The Difference between GANC and GGC}
\label{sec:diff}
Although the workflows of GANC and GGC are similar, they are essentially different in many aspects.
1. GANC is essentially a hierarchical clustering algorithm. 
It first performs complete hierarchical clustering, 
then uses a refinement step to reassign boundary samples to obtain the final clusters.
Meanwhile, GGC aims to optimize a global loss function,
therefore the final partition corresponding to a lower loss will be obtained directly,
which results in that GGC often yields more superior experimental performance
2. GGC and GANC adopt different acceleration technologies, resulting in different time complexities.
Especially, GANC's time complexity is $O(n_g \log^2(n))$, 
where $n_g$ is the number of edges in the graph, 
and GGC's complexity is only $O((n-c)k_1\log(nk_1))$. 
Since $n_g \geq n k_1$, we have 
\begin{equation*}
  \frac{\text{Time Complexity of GANC}}{\text{Time Complexity of GGC}} = \frac{n_g \log^2 (n)}{(n-c)k_1 \log(n)} = \eta \log(n),
\end{equation*}
where $\eta > 1$ is a constant.
Therefore, GGC's complexity is much lower. 
This results in GGC having advantages in running time, especially on large-scale data.
These differences are confirmed by experimental results.




\section{Experiments}

\subsection{Datasets}
16 middle-scale (Face96, Yale64, JAFFE, CNBC, Digits, PINS, Srbct, Madelon, USPS, Grimace, Face94, Palm, FaceV5, IMM, Coil20, 
and USPST) and 
7 large-scale (PEAL, Emnist, Eletters, Ebalanced, CACD, CelebA, and YouTubeFace) real-world datasets are used to 
verify the performance of GGC.
To save table width, we denote the middle-scale datasets as $M_1$-$M_{16}$, and the large-scale datasets as $L_1$-$L_7$.
The number of samples these datasets contained is presented in the Table \ref{tab:mid} and Table \ref{tab:large}.
For these face datasets, we adopt the method proposed in \cite{ksums} to extract the features.

\subsection{Baselines}
We choose the following algorithms as baselines: 
the Traditional Spectral Clustering algorithm (TSC) \cite{ncut2},
Ultra-Scalable spectral clustering and Ensemble Clustering (USEC) \cite{uspec},
Fast Clustering with co-clustering via Discrete non-negative Matrix Factorization (FCDMF) \cite{fcdmf},
fast optimization of Spectral Embedding and Improved Spectral Rotation (SE-ISR) \cite{sesr}, and
Improved Anchor-based Graph Clustering (IAGC) \cite{AGCI}.
Furthermore, Greedy Agglomerative Normalized Cut (GANC) \cite{GANC} is also included as a baseline.
Although GANC is a method for hierarchical clustering, its merge operation is similar to the proposed algorithm. 

The parameters involved in these algorithms are described below. 
For algorithms that take a $k_2$-NN graph or an anchor graph as input, we use the method proposed in \cite{CLR} 
to compute the similarities between nodes in the graph. 
The number of neighbors involved in both the $k_2$-NN graph and anchor graph is determined as follows: 
\begin{equation}
k_2 = \min\{50, \lfloor n/c \rfloor \},
\end{equation} 
where $n$ is the number of samples and $c$ is the number of clusters. 
For the number of anchors, $a$, involved in anchor graph, we fix it as $a=1024$.
For $k$-means involved in the above algorithms, the version proposed in \cite{kmpp} is adopted.
The algorithms were run 50 times repeatedly and the average performance is reported.

All algorithms are run on an Arch machine with an Intel(R) Core(TM) i7-10700K CPU @ 3.8GHz and 32 GB of main memory.

\begin{table*}[]
\renewcommand{\arraystretch}{0.95}
\caption{The performance of the algorithms on the middle-scale real-world datasets. The best results are shown in bold.}
\label{tab:mid}
\centering
\begin{threeparttable}
  \begin{tabular}{ccccccccc}
  \toprule
  Datasets                  & Metrics  & TSC                         & USPEC                       & FCDMF                       & SESR                        & IAGC                        & GANC \footnotemark[1]          & GGC            \\ \midrule
  \multirow{4}{*}{\makecell{$M_{1}$ \\ 3016}}  & ACC      & 0.877 ($\pm$0.005)          & 0.965 ($\pm$0.003)          & 0.567 ($\pm$0.032)          & 0.966 ($\pm$0.010)          & 0.955 ($\pm$0.004)          & 0.969          & \textbf{0.973} \\
                            & NMI      & 0.924 ($\pm$0.001)          & 0.983 ($\pm$0.001)          & 0.733 ($\pm$0.034)          & 0.976 ($\pm$0.002)          & 0.979 ($\pm$0.001)          & 0.987          & \textbf{0.988} \\
                            & ARI      & 0.561 ($\pm$0.015)          & 0.943 ($\pm$0.004)          & 0.099 ($\pm$0.031)          & 0.883 ($\pm$0.006)          & 0.934 ($\pm$0.004)          & 0.936          & \textbf{0.949} \\
                            & Time (s) & 0.384 ($\pm$0.001)          & 0.085 ($\pm$0.004)          & 0.425 ($\pm$0.002)          & 0.017 ($\pm$0.124)          & 1.073 ($\pm$0.005)          & 0.336          & \textbf{0.003} \\ \midrule
  \multirow{4}{*}{\makecell{$M_{2}$ \\ 165}}  & ACC      & 0.605 ($\pm$0.018)          & 0.581 ($\pm$0.018)          & 0.509 ($\pm$0.063)          & 0.551 ($\pm$0.012)          & \textbf{0.606 ($\pm$0.012)} & 0.461          & 0.570          \\
                            & NMI      & \textbf{0.648 ($\pm$0.008)} & 0.622 ($\pm$0.010)          & 0.546 ($\pm$0.050)          & 0.560 ($\pm$0.015)          & 0.609 ($\pm$0.010)          & 0.550          & 0.621          \\
                            & ARI      & \textbf{0.422 ($\pm$0.010)} & 0.391 ($\pm$0.013)          & 0.279 ($\pm$0.066)          & 0.303 ($\pm$0.021)          & 0.356 ($\pm$0.015)          & 0.311          & 0.381          \\
                            & Time (s) & 0.012 ($\pm$0.000)          & 0.012 ($\pm$0.003)          & 0.001 ($\pm$0.000)          & 0.000 ($\pm$0.003)          & 0.014 ($\pm$0.000)          & 0.022          & \textbf{0.000} \\ \midrule
  \multirow{4}{*}{\makecell{$M_{3}$ \\ 213}}  & ACC      & 0.862 ($\pm$0.003)          & \textbf{1.000 ($\pm$0.000)} & \textbf{0.586 ($\pm$0.112)} & \textbf{1.000 ($\pm$0.000)} & \textbf{1.000 ($\pm$0.000)} & \textbf{1.000} & \textbf{1.000} \\
                            & NMI      & 0.939 ($\pm$0.000)          & \textbf{1.000 ($\pm$0.000)} & \textbf{0.732 ($\pm$0.105)} & \textbf{1.000 ($\pm$0.000)} & \textbf{1.000 ($\pm$0.000)} & \textbf{1.000} & \textbf{1.000} \\
                            & ARI      & 0.862 ($\pm$0.002)          & \textbf{1.000 ($\pm$0.000)} & \textbf{0.473 ($\pm$0.156)} & \textbf{1.000 ($\pm$0.000)} & \textbf{1.000 ($\pm$0.000)} & \textbf{1.000} & \textbf{1.000} \\
                            & Time (s) & 0.007 ($\pm$0.000)          & 0.013 ($\pm$0.000)          & 0.001 ($\pm$0.000)          & \textbf{0.000 ($\pm$0.002)} & 0.016 ($\pm$0.000)          & 0.030          & \textbf{0.000} \\ \midrule
  \multirow{4}{*}{\makecell{$M_{4}$ \\ 6574}}  & ACC      & \textbf{0.660 ($\pm$0.003)} & 0.609 ($\pm$0.003)          & 0.440 ($\pm$0.011)          & 0.620 ($\pm$0.003)          & 0.616 ($\pm$0.002)          & 0.347          & 0.644          \\
                            & NMI      & \textbf{0.777 ($\pm$0.003)} & 0.753 ($\pm$0.002)          & 0.663 ($\pm$0.015)          & 0.701 ($\pm$0.004)          & 0.764 ($\pm$0.002)          & 0.526          & 0.704          \\
                            & ARI      & 0.329 ($\pm$0.019)          & 0.387 ($\pm$0.006)          & 0.115 ($\pm$0.029)          & 0.149 ($\pm$0.022)          & \textbf{0.397 ($\pm$0.008)} & 0.046          & 0.056          \\
                            & Time (s) & 19.272 ($\pm$0.022)         & 0.328 ($\pm$0.014)          & 1.353 ($\pm$0.008)          & 0.136 ($\pm$0.968)          & 5.535 ($\pm$0.036)          & 0.713          & \textbf{0.011} \\ \midrule
  \multirow{4}{*}{\makecell{$M_{5}$ \\ 4000}}  & ACC      & 0.646 ($\pm$0.012)          & 0.457 ($\pm$0.003)          & 0.444 ($\pm$0.019)          & 0.482 ($\pm$0.008)          & 0.460 ($\pm$0.006)          & 0.583          & \textbf{0.717} \\
                            & NMI      & 0.649 ($\pm$0.005)          & 0.470 ($\pm$0.004)          & 0.488 ($\pm$0.021)          & 0.499 ($\pm$0.003)          & 0.486 ($\pm$0.004)          & 0.538          & \textbf{0.708} \\
                            & ARI      & 0.528 ($\pm$0.007)          & 0.342 ($\pm$0.007)          & 0.341 ($\pm$0.034)          & 0.374 ($\pm$0.007)          & 0.339 ($\pm$0.004)          & 0.498          & \textbf{0.597} \\
                            & Time (s) & 0.053 ($\pm$0.001)          & 0.062 ($\pm$0.001)          & 0.281 ($\pm$0.002)          & 0.023 ($\pm$0.172)          & 1.000 ($\pm$0.002)          & 0.492          & \textbf{0.022} \\ \midrule
  \multirow{4}{*}{\makecell{$M_{6}$ \\ 10770}}  & ACC      & 0.942 ($\pm$0.009)          & \textbf{0.959 ($\pm$0.007)} & 0.620 ($\pm$0.025)          & 0.959 ($\pm$0.003)          & 0.937 ($\pm$0.004)          & 0.119          & 0.950          \\
                            & NMI      & 0.967 ($\pm$0.003)          & \textbf{0.974 ($\pm$0.001)} & 0.777 ($\pm$0.022)          & 0.956 ($\pm$0.001)          & 0.955 ($\pm$0.002)          & 0.128          & 0.956          \\
                            & ARI      & 0.919 ($\pm$0.018)          & \textbf{0.948 ($\pm$0.008)} & 0.238 ($\pm$0.059)          & 0.859 ($\pm$0.007)          & 0.892 ($\pm$0.015)          & 0.005          & 0.841          \\
                            & Time (s) & 2.316 ($\pm$0.002)          & 0.128 ($\pm$0.011)          & 1.784 ($\pm$0.010)          & 0.062 ($\pm$0.453)          & 3.963 ($\pm$0.013)          & 1.400          & \textbf{0.032} \\ \midrule
  \multirow{4}{*}{\makecell{$M_{7}$ \\ 63}}  & ACC      & 0.649 ($\pm$0.005)          & 0.573 ($\pm$0.015)          & 0.605 ($\pm$0.071)          & 0.571 ($\pm$0.000)          & \textbf{0.651 ($\pm$0.017)} & 0.556          & 0.619          \\
                            & NMI      & 0.442 ($\pm$0.010)          & 0.340 ($\pm$0.014)          & 0.401 ($\pm$0.046)          & 0.343 ($\pm$0.000)          & 0.454 ($\pm$0.030)          & 0.392          & \textbf{0.486} \\
                            & ARI      & 0.278 ($\pm$0.008)          & 0.164 ($\pm$0.021)          & 0.234 ($\pm$0.045)          & 0.157 ($\pm$0.000)          & \textbf{0.293 ($\pm$0.030)} & 0.254          & 0.248          \\
                            & Time (s) & 0.007 ($\pm$0.000)          & 0.006 ($\pm$0.000)          & 0.000 ($\pm$0.000)          & \textbf{0.000 ($\pm$0.000)} & 0.009 ($\pm$0.000)          & 0.013          & \textbf{0.000} \\ \midrule
  \multirow{4}{*}{\makecell{$M_{8}$ \\ 2600}}  & ACC      & 0.501 ($\pm$0.000)          & 0.501 ($\pm$0.000)          & 0.520 ($\pm$0.000)          & 0.504 ($\pm$0.002)          & 0.519 ($\pm$0.000)          & 0.501          & \textbf{0.540} \\
                            & NMI      & 0.000 ($\pm$0.000)          & 0.000 ($\pm$0.000)          & 0.001 ($\pm$0.000)          & 0.000 ($\pm$0.000)          & 0.001 ($\pm$0.000)          & 0.000          & \textbf{0.005} \\
                            & ARI      & 0.000 ($\pm$0.000)          & 0.000 ($\pm$0.000)          & 0.001 ($\pm$0.000)          & 0.000 ($\pm$0.000)          & 0.001 ($\pm$0.000)          & 0.000          & \textbf{0.006} \\
                            & Time (s) & 0.016 ($\pm$0.000)          & 0.046 ($\pm$0.001)          & 0.207 ($\pm$0.015)          & \textbf{0.006 ($\pm$0.044)} & 1.013 ($\pm$0.001)          & 0.347          & 0.011          \\ \midrule
  \multirow{4}{*}{\makecell{$M_{9}$ \\ 9298}}  & ACC      & 0.694 ($\pm$0.051)          & 0.659 ($\pm$0.049)          & 0.627 ($\pm$0.079)          & 0.625 ($\pm$0.009)          & 0.505 ($\pm$0.023)          & 0.167          & \textbf{0.838} \\
                            & NMI      & 0.831 ($\pm$0.010)          & 0.672 ($\pm$0.012)          & 0.653 ($\pm$0.047)          & 0.694 ($\pm$0.004)          & 0.612 ($\pm$0.009)          & 0.000          & \textbf{0.867} \\
                            & ARI      & 0.701 ($\pm$0.035)          & 0.580 ($\pm$0.038)          & 0.502 ($\pm$0.084)          & 0.551 ($\pm$0.004)          & 0.411 ($\pm$0.014)          & 0.000          & \textbf{0.824} \\
                            & Time (s) & 0.199 ($\pm$0.000)          & 0.079 ($\pm$0.004)          & 0.722 ($\pm$0.013)          & \textbf{0.042 ($\pm$0.306)} & 2.240 ($\pm$0.002)          & 1.243          & 0.058          \\ \midrule
  \multirow{4}{*}{\makecell{$M_{10}$ \\ 360}} & ACC      & 0.721 ($\pm$0.028)          & \textbf{0.981 ($\pm$0.000)} & 0.613 ($\pm$0.117)          & 0.971 ($\pm$0.028)          & 0.949 ($\pm$0.025)          & 0.936          & 0.981          \\
                            & NMI      & 0.872 ($\pm$0.017)          & \textbf{0.985 ($\pm$0.000)} & 0.762 ($\pm$0.098)          & 0.979 ($\pm$0.009)          & 0.969 ($\pm$0.010)          & 0.961          & 0.985          \\
                            & ARI      & 0.681 ($\pm$0.035)          & \textbf{0.965 ($\pm$0.000)} & 0.444 ($\pm$0.167)          & 0.953 ($\pm$0.024)          & 0.930 ($\pm$0.024)          & 0.909          & \textbf{0.965} \\
                            & Time (s) & 0.008 ($\pm$0.000)          & 0.018 ($\pm$0.005)          & 0.004 ($\pm$0.000)          & \textbf{0.000 ($\pm$0.004)} & 0.026 ($\pm$0.000)          & 0.049          & 0.001          \\ \midrule
  \multirow{4}{*}{\makecell{$M_{11}$ \\ 2640}} & ACC      & 0.773 ($\pm$0.007)          & 0.899 ($\pm$0.012)          & 0.578 ($\pm$0.036)          & 0.932 ($\pm$0.010)          & 0.926 ($\pm$0.006)          & 0.957          & \textbf{0.973} \\
                            & NMI      & 0.916 ($\pm$0.003)          & 0.968 ($\pm$0.002)          & 0.751 ($\pm$0.030)          & 0.979 ($\pm$0.002)          & 0.979 ($\pm$0.001)          & 0.989          & \textbf{0.993} \\
                            & ARI      & 0.615 ($\pm$0.025)          & 0.883 ($\pm$0.010)          & 0.107 ($\pm$0.028)          & 0.922 ($\pm$0.009)          & 0.919 ($\pm$0.005)          & 0.958          & \textbf{0.972} \\
                            & Time (s) & 0.192 ($\pm$0.001)          & 0.101 ($\pm$0.002)          & 0.275 ($\pm$0.003)          & 0.020 ($\pm$0.142)          & 0.908 ($\pm$0.003)          & 0.296          & \textbf{0.003} \\ \midrule
  \multirow{4}{*}{\makecell{$M_{12}$ \\ 2000}} & ACC      & \textbf{0.882 ($\pm$0.005)} & 0.808 ($\pm$0.016)          & 0.570 ($\pm$0.040)          & 0.830 ($\pm$0.006)          & 0.833 ($\pm$0.010)          & 0.862          & 0.872          \\
                            & NMI      & \textbf{0.965 ($\pm$0.001)} & 0.926 ($\pm$0.004)          & 0.776 ($\pm$0.037)          & 0.944 ($\pm$0.002)          & 0.939 ($\pm$0.002)          & 0.961          & 0.964          \\
                            & ARI      & \textbf{0.873 ($\pm$0.004)} & 0.767 ($\pm$0.013)          & 0.200 ($\pm$0.081)          & 0.801 ($\pm$0.007)          & 0.789 ($\pm$0.010)          & 0.851          & 0.863          \\
                            & Time (s) & 0.145 ($\pm$0.000)          & 0.080 ($\pm$0.002)          & 0.288 ($\pm$0.004)          & 0.015 ($\pm$0.105)          & 0.544 ($\pm$0.003)          & 0.250          & \textbf{0.003} \\ \midrule
  \multirow{4}{*}{\makecell{$M_{13}$ \\ 2500}} & ACC      & 0.894 ($\pm$0.003)          & 0.885 ($\pm$0.005)          & 0.575 ($\pm$0.016)          & 0.934 ($\pm$0.005)          & 0.917 ($\pm$0.003)          & 0.940          & \textbf{0.970} \\
                            & NMI      & 0.955 ($\pm$0.002)          & 0.963 ($\pm$0.001)          & 0.738 ($\pm$0.014)          & 0.976 ($\pm$0.001)          & 0.973 ($\pm$0.001)          & 0.980          & \textbf{0.987} \\
                            & ARI      & 0.621 ($\pm$0.039)          & 0.815 ($\pm$0.006)          & 0.024 ($\pm$0.003)          & 0.791 ($\pm$0.035)          & 0.740 ($\pm$0.021)          & 0.884          & \textbf{0.885} \\
                            & Time (s) & 8.058 ($\pm$0.013)          & 0.245 ($\pm$0.014)          & 0.643 ($\pm$0.007)          & 0.096 ($\pm$0.681)          & 4.180 ($\pm$0.016)          & 0.255          & \textbf{0.001} \\ \midrule
  \multirow{4}{*}{\makecell{$M_{14}$ \\ 240}} & ACC      & 0.628 ($\pm$0.011)          & 0.603 ($\pm$0.008)          & 0.500 ($\pm$0.047)          & 0.575 ($\pm$0.008)          & 0.624 ($\pm$0.008)          & 0.471          & \textbf{0.633} \\
                            & NMI      & \textbf{0.794 ($\pm$0.005)} & 0.787 ($\pm$0.004)          & 0.684 ($\pm$0.029)          & 0.768 ($\pm$0.005)          & 0.762 ($\pm$0.005)          & 0.648          & 0.761          \\
                            & ARI      & \textbf{0.446 ($\pm$0.010)} & 0.439 ($\pm$0.010)          & 0.223 ($\pm$0.054)          & 0.406 ($\pm$0.010)          & 0.370 ($\pm$0.018)          & 0.234          & 0.387          \\
                            & Time (s) & 0.039 ($\pm$0.001)          & 0.013 ($\pm$0.001)          & 0.003 ($\pm$0.000)          & 0.001 ($\pm$0.008)          & 0.069 ($\pm$0.126)          & 0.032          & \textbf{0.000} \\ \midrule
  \multirow{4}{*}{\makecell{$M_{15}$ \\ 1440}} & ACC      & 0.777 ($\pm$0.018)          & 0.733 ($\pm$0.014)          & 0.558 ($\pm$0.072)          & 0.741 ($\pm$0.015)          & 0.684 ($\pm$0.006)          & 0.812          & \textbf{0.824} \\
                            & NMI      & 0.866 ($\pm$0.010)          & 0.805 ($\pm$0.006)          & 0.721 ($\pm$0.041)          & 0.811 ($\pm$0.005)          & 0.794 ($\pm$0.004)          & 0.899          & \textbf{0.909} \\
                            & ARI      & 0.740 ($\pm$0.020)          & 0.674 ($\pm$0.012)          & 0.500 ($\pm$0.083)          & 0.664 ($\pm$0.012)          & 0.620 ($\pm$0.007)          & 0.719          & \textbf{0.744} \\
                            & Time (s) & 0.040 ($\pm$0.000)          & 0.067 ($\pm$0.000)          & 0.050 ($\pm$0.000)          & 0.006 ($\pm$0.042)          & 0.164 ($\pm$0.001)          & 0.167          & \textbf{0.005} \\ \midrule
  \multirow{4}{*}{\makecell{$M_{16}$ \\ 2007}} & ACC      & 0.710 ($\pm$0.053)          & 0.625 ($\pm$0.046)          & 0.642 ($\pm$0.033)          & 0.626 ($\pm$0.005)          & 0.609 ($\pm$0.026)          & 0.709          & \textbf{0.718} \\
                            & NMI      & 0.787 ($\pm$0.017)          & 0.652 ($\pm$0.015)          & 0.667 ($\pm$0.019)          & 0.685 ($\pm$0.001)          & 0.669 ($\pm$0.007)          & \textbf{0.806} & 0.775          \\
                            & ARI      & 0.667 ($\pm$0.049)          & 0.534 ($\pm$0.037)          & 0.527 ($\pm$0.031)          & 0.538 ($\pm$0.003)          & 0.508 ($\pm$0.015)          & 0.680          & \textbf{0.695} \\
                            & Time (s) & 0.025 ($\pm$0.000)          & 0.052 ($\pm$0.000)          & 0.121 ($\pm$0.006)          & \textbf{0.006 ($\pm$0.041)} & 0.205 ($\pm$0.001)          & 0.270          & 0.009          \\ \bottomrule
  \end{tabular}
  \begin{tablenotes}
    \item[1] https://www.tsp.ece.mcgill.ca/Networks/projects/proj-ganc.html
  \end{tablenotes}
\end{threeparttable}
\end{table*}

\begin{table*}[]
  \centering
  \begin{threeparttable}[t]
\renewcommand{\arraystretch}{1.0}
\caption{The performance of the algorithms on the large-scale real-world datasets \tnote{1}. The best results are shown in bold.}
\label{tab:large}
  \begin{tabular}{ccccccccc}
  \toprule
  Data                                          & Metrics & TSC                         & USPEC                       & FCDMF                       & SESR                        & IAGC                   & GANC    & GGC            \\ \midrule
  \multirow{4}{*}{\makecell{$L_{1}$ \\ 30863}}  & ACC     & \textbf{0.912 ($\pm$0.001)} & 0.615 ($\pm$0.005)          & 0.587 ($\pm$0.010)          & 0.630 ($\pm$0.006)          & 0.609 ($\pm$0.006)     & 0.798   & 0.890          \\
                                                & NMI     & \textbf{0.951 ($\pm$0.000)} & 0.815 ($\pm$0.001)          & 0.818 ($\pm$0.006)          & 0.800 ($\pm$0.006)          & 0.818 ($\pm$0.001)     & 0.856   & 0.913          \\
                                                & ARI     & \textbf{0.723 ($\pm$0.007)} & 0.289 ($\pm$0.004)          & 0.155 ($\pm$0.028)          & 0.054 ($\pm$0.011)          & 0.378 ($\pm$0.016)     & 0.090   & 0.168          \\
                                                & Time(s) & 1343.163 ($\pm$1.901)       & 19.810 ($\pm$0.826)         & 11.647 ($\pm$0.349)         & 361.248 ($\pm$54.960)       & 138.055 ($\pm$2.438)   & 3.973   & \textbf{0.111} \\ \midrule
  \multirow{4}{*}{\makecell{$L_{2}$ \\ 60000}}  & ACC     & 0.775 ($\pm$0.038)          & 0.505 ($\pm$0.034)          & 0.532 ($\pm$0.014)          & 0.522 ($\pm$0.008)          & 0.533 ($\pm$0.009)     & 0.100   & \textbf{0.823} \\
                                                & NMI     & 0.785 ($\pm$0.011)          & 0.421 ($\pm$0.019)          & 0.521 ($\pm$0.004)          & 0.492 ($\pm$0.005)          & 0.494 ($\pm$0.004)     & 0.000   & \textbf{0.858} \\
                                                & ARI     & 0.692 ($\pm$0.023)          & 0.305 ($\pm$0.027)          & 0.387 ($\pm$0.008)          & 0.345 ($\pm$0.007)          & 0.360 ($\pm$0.002)     & 0.000   & \textbf{0.808} \\
                                                & Time(s) & 5.510 ($\pm$0.009)          & 2.640 ($\pm$0.050)          & 6.137 ($\pm$0.485)          & 13.142 ($\pm$4.122)         & 10.560 ($\pm$0.005)    & 31.572  & 3.260          \\ \midrule
  \multirow{4}{*}{\makecell{$L_{3}$ \\ 88800}}  & ACC     & 0.549 ($\pm$0.010)          & 0.313 ($\pm$0.008)          & 0.386 ($\pm$0.007)          & 0.361 ($\pm$0.013)          & 0.296 ($\pm$0.006)     & 0.039   & \textbf{0.559} \\
                                                & NMI     & 0.624 ($\pm$0.004)          & 0.342 ($\pm$0.003)          & 0.408 ($\pm$0.003)          & 0.394 ($\pm$0.004)          & 0.346 ($\pm$0.002)     & 0.000   & \textbf{0.626} \\
                                                & ARI     & 0.440 ($\pm$0.006)          & 0.162 ($\pm$0.003)          & 0.223 ($\pm$0.003)          & 0.202 ($\pm$0.006)          & 0.159 ($\pm$0.003)     & 0.000   & \textbf{0.453} \\
                                                & Time(s) & 15.176 ($\pm$0.025)         & 4.480 ($\pm$0.077)          & 8.862 ($\pm$0.380)          & 36.419 ($\pm$7.444)         & 16.971 ($\pm$0.022)    & 44.924  & 6.216          \\ \midrule
  \multirow{4}{*}{\makecell{$L_{4}$ \\ 112800}} & ACC     & \textbf{0.526 ($\pm$0.009)} & 0.266 ($\pm$0.002)          & 0.331 ($\pm$0.007)          & 0.306 ($\pm$0.009)          & 0.256 ($\pm$0.005)     & 0.021   & 0.521          \\
                                                & NMI     & \textbf{0.634 ($\pm$0.002)} & 0.360 ($\pm$0.001)          & 0.428 ($\pm$0.003)          & 0.407 ($\pm$0.004)          & 0.360 ($\pm$0.002)     & 0.000   & 0.621          \\
                                                & ARI     & \textbf{0.392 ($\pm$0.004)} & 0.136 ($\pm$0.001)          & 0.184 ($\pm$0.003)          & 0.164 ($\pm$0.004)          & 0.131 ($\pm$0.002)     & 0.000   & 0.374          \\
                                                & Time(s) & 28.267 ($\pm$0.060)         & 6.786 ($\pm$0.176)          & 12.259 ($\pm$0.137)         & 63.396 ($\pm$11.511)        & 28.471 ($\pm$0.141)    & 63.146  & 9.654          \\ \midrule
  \multirow{4}{*}{\makecell{$L_{5}$ \\ 163446}} & ACC     & -                           & 0.520 ($\pm$0.003)          & \textbf{0.589 ($\pm$0.007)} & -                           & 0.018 ($\pm$0.000)     & 0.117   & 0.469          \\
                                                & NMI     & -                           & 0.755 ($\pm$0.001)          & \textbf{0.783 ($\pm$0.005)} & -                           & 0.283 ($\pm$0.004)     & 0.150   & 0.575          \\
                                                & ARI     & -                           & \textbf{0.359 ($\pm$0.002)} & 0.043 ($\pm$0.005)          & -                           & 0.001 ($\pm$0.000)     & 0.000   & 0.004          \\
                                                & Time(s) & -                           & 596.841 ($\pm$84.697)       & 186.017 ($\pm$3.240)        & -                           & 2168.368 ($\pm$74.606) & 30.105  & \textbf{1.946} \\ \midrule
  \multirow{4}{*}{\makecell{$L_{6}$ \\ 202599}} & ACC     & -                           & -                           & -                           & -                           & -                      & 0.246   & \textbf{0.480} \\
                                                & NMI     & -                           & -                           & -                           & -                           & -                      & 0.394   & \textbf{0.592} \\
                                                & ARI     & -                           & -                           & -                           & -                           & -                      & 0.001   & \textbf{0.002} \\
                                                & Time(s) & -                           & -                           & -                           & -                           & -                      & 18.688  & \textbf{0.296} \\ \midrule
  \multirow{4}{*}{\makecell{$L_{7}$ \\ 621126}} & ACC     & -                           & -                           & -                           & -                           & -                      & 0.661   & \textbf{0.701} \\
                                                & NMI     & -                           & -                           & -                           & -                           & -                      & 0.813   & \textbf{0.854} \\
                                                & ARI     & -                           & -                           & -                           & -                           & -                      & 0.070   & \textbf{0.092} \\
                                                & Time(s) & -                           & -                           & -                           & -                           & -                      & 235.815 & \textbf{32.64} \\ \midrule
  \end{tabular}
  \begin{tablenotes}
    \item[1] ``-'' is used to denote cases where the algorithm failed due to insufficient RAM.
  \end{tablenotes}
  \end{threeparttable}
\end{table*}

\subsection{Metrics}
Three metrics are adopted to measure the performance of the proposed algorithm, including ACCuracy (ACC), Adjusted Rand Index (ARI) and Normalized Mutual Information (NMI). They are calculated as follows:
\begin{equation}
NMI(\mathcal{A}, \mathcal{B}) = \frac{ 
\sum_{i=1}^c \sum_{j=1}^c \frac{ \vert \mathcal{A}_i \cap \mathcal{B}_j \vert }{n}
\log \frac{ 
          n \vert \mathcal{A}_i \cap \mathcal{B}_j \vert
     }{
          \vert \mathcal{A}_i \vert  \vert \mathcal{B}_j \vert
     }
}{
\sum_{i=1}^c \left( 
                   \frac{\vert \mathcal{A}_i \vert}{n} \log \frac{\vert \mathcal{A}_i \vert}{n} +  
                   \frac{\vert \mathcal{B}_i \vert}{n} \log \frac{\vert \mathcal{B}_i \vert}{n} 
             \right)
},
\end{equation}
\begin{equation}
ARI(\mathcal{A}, \mathcal{B}) = \frac{
\sum_{i,j=1}^c  
\begin{pmatrix}
\vert \mathcal{A}_i \cap \mathcal{B}_j \vert \\
2
\end{pmatrix} - E[RI]
}{
\frac{1}{2}\sum_{i=1}^c \left(
\begin{pmatrix}
\vert \mathcal{A}_i \vert \\
2
\end{pmatrix} + 
\begin{pmatrix}
\vert \mathcal{B}_i \vert \\
2
\end{pmatrix}  \right) - E[RI]
},
\end{equation}
\begin{equation}
ACC(\y, \hat{\y}) = \sum_{i=1}^n \delta ( map(\hat{y}_i, y_i) ) / n,
\end{equation}
where $\y$ and $\hat{\y}$ are the ground truth and predicted labels, respectively, $map$ is a function to find the best match
between cluster label and true label,
$\mathcal{A}_i = \{ \x_j \mid y_j = i \}$,
$\mathcal{B}_i = \{ \x_j \mid \hat{y}_j = i \}$, and
\begin{equation}
E[RI] =  \left. 
\sum_{i=1}^c 
\begin{pmatrix}
\vert \mathcal{A}_i \vert \\
2
\end{pmatrix}
\sum_{j=1}^c 
\begin{pmatrix}
\vert \mathcal{B}_j \vert \\
2
\end{pmatrix} \middle/ 
\begin{pmatrix}
n \\
2
\end{pmatrix} \right. .
\end{equation} 

\subsection{Experimental Results on Real-world Datasets}
\label{sec:exp}
From the experimental results shown in 
Table \ref{tab:obj}, Table \ref{tab:mid}, and Table \ref{tab:large}, we can see that: 
\begin{itemize}
\item TSC and GGC share the same loss function,
so, it is meaningful to compare their losses. 
Table \ref{tab:obj} shows that GGC's loss is smaller than TSC's in most cases,
demonstrating the superiority of the proposed strategy over the classical two-step approach.
\item Table \ref{tab:mid} shows GGC achieves the best performance in most cases.
Assume that the first place gets 6 points, the second place gets 5 point, and so on.
Then, the total scores are: TSC-233, USPEC-201, FCDMF-104, SESR-164, IAGC-233, GANC-214, GGC-255 on middle-scale datasets,
which verifies the effectiveness of the proposed algorithm.
\item The computational complexity of GGC is nearly linear w.r.t the number of samples, 
which results in it always having a shorter running time.
The more samples in the dataset, the more prominent GGC's efficiency advantage is.
Specifically, for datasets with less than 100k samples, 
the running time difference between these fast algorithms is small. 
For datasets with more than 100k samples, 
GGC's efficiency advantage becomes more prominent. 
For example, 
On CACD ($L_5$), GGC saves 94\% of time compared to GANC.
For datasets with over 200k samples, 
anchor-based fast algorithms like USPEC cannot run due to the need for large memory, 
which is a common dilemma for anchor-based algorithms,
while GGC is more memory efficient and can scale to larger datasets.
\item There is a significant difference between GANC and GGC in terms of clustering quality, 
and running time, 
as shown in Table \ref{tab:mid}, and Table \ref{tab:large},
which confirms what was discussed in Section \ref{sec:diff}
\end{itemize}

\section{Conclusion}
We propose a new optimization method (GGC) for the normalized cut problem where no hyperparameters are involved,
and the final partition obtained is unique.
Furthermore, the GGC algorithm has an approximately linear time complexity in terms of the number of samples, 
enabling easy grouping of large-scale datasets.
The results of experiments conducted on 16 middle-scale and 7 large-scale real-world datasets 
demonstrate the superiority of the proposed algorithm.
As a new optimization method for the normalized cut problem, 
we compare the loss function value obtained by GGC and the traditional optimization method, 
finding GGC achieves lower loss in most cases.
It is worth mentioning that the proposed strategy can also be used to solve other problems besides the normalized cut, 
such as Ratio-cut and KSUMS. 





%

\ifCLASSOPTIONcompsoc
  \section*{Acknowledgments}
  This work was supported in part by the National Natural Science Foundation of China under Grant 62236001, Grant 62276212 and Grant 62176212.
\else
  \section*{Acknowledgment}
\fi


\ifCLASSOPTIONcaptionsoff
  \newpage
\fi



\bibliographystyle{IEEEtran}
\bibliography{my}

\begin{thebibliography}{10}
\providecommand{\url}[1]{#1}
\csname url@samestyle\endcsname
\providecommand{\newblock}{\relax}
\providecommand{\bibinfo}[2]{#2}
\providecommand{\BIBentrySTDinterwordspacing}{\spaceskip=0pt\relax}
\providecommand{\BIBentryALTinterwordstretchfactor}{4}
\providecommand{\BIBentryALTinterwordspacing}{\spaceskip=\fontdimen2\font plus
\BIBentryALTinterwordstretchfactor\fontdimen3\font minus \fontdimen4\font\relax}
\providecommand{\BIBforeignlanguage}[2]{{%
\expandafter\ifx\csname l@#1\endcsname\relax
\typeout{** WARNING: IEEEtran.bst: No hyphenation pattern has been}%
\typeout{** loaded for the language `#1'. Using the pattern for}%
\typeout{** the default language instead.}%
\else
\language=\csname l@#1\endcsname
\fi
#2}}
\providecommand{\BIBdecl}{\relax}
\BIBdecl

\bibitem{cc}
S.~Pei, H.~Chen, F.~Nie, R.~Wang, and X.~Li, ``Centerless clustering: An efficient variant of k-means based on k-nn graph,'' \emph{IEEE Transactions on Pattern Analysis and Machine Intelligence}, 2022.

\bibitem{cluster_text}
A.~Raj and S.~Susan, ``Clustering analysis for newsgroup classification,'' in \emph{Data Engineering and Intelligent Computing}.\hskip 1em plus 0.5em minus 0.4em\relax Springer, 2022, pp. 271--279.

\bibitem{cluster_face}
C.~Otto, D.~Wang, and A.~K. Jain, ``Clustering millions of faces by identity,'' \emph{IEEE transactions on pattern analysis and machine intelligence}, vol.~40, no.~2, pp. 289--303, 2017.

\bibitem{cluster_seg}
H.~Mittal, A.~C. Pandey, M.~Saraswat, S.~Kumar, R.~Pal, and G.~Modwel, ``A comprehensive survey of image segmentation: clustering methods, performance parameters, and benchmark datasets,'' \emph{Multimedia Tools and Applications}, pp. 1--26, 2021.

\bibitem{ncut2}
A.~Y. Ng, M.~I. Jordan, and Y.~Weiss, ``On spectral clustering: Analysis and an algorithm,'' in \emph{Proceedings of the 14th International Conference on Neural Information Processing Systems: Natural and Synthetic}, ser. NIPS'01.\hskip 1em plus 0.5em minus 0.4em\relax Cambridge, MA, USA: MIT Press, 2001, p. 849–856.

\bibitem{cluster_tur}
U.~Von~Luxburg, ``A tutorial on spectral clustering,'' \emph{Statistics and computing}, vol.~17, no.~4, pp. 395--416, 2007.

\bibitem{kmeans}
S.~Lloyd, ``Least squares quantization in pcm,'' \emph{IEEE Transactions on Information Theory}, vol.~28, no.~2, pp. 129--137, 1982.

\bibitem{SR}
S.~X. Yu and J.~Shi, ``Multiclass spectral clustering,'' in \emph{Proceedings Ninth IEEE International Conference on Computer Vision}, 2003, pp. 313--319 vol.1.

\bibitem{rcut}
L.~Hagen and A.~Kahng, ``New spectral methods for ratio cut partitioning and clustering,'' \emph{IEEE Transactions on Computer-Aided Design of Integrated Circuits and Systems}, vol.~11, no.~9, pp. 1074--1085, 1992.

\bibitem{ISR}
X.~Chen, F.~Nie, J.~Z. Huang, and M.~Yang, ``Scalable normalized cut with improved spectral rotation,'' in \emph{Proceedings of the 26th International Joint Conference on Artificial Intelligence}, ser. IJCAI'17.\hskip 1em plus 0.5em minus 0.4em\relax AAAI Press, 2017, p. 1518–1524.

\bibitem{mpsc}
S.~Ding, L.~Cong, Q.~Hu, H.~Jia, and Z.~Shi, ``A multiway p-spectral clustering algorithm,'' \emph{Knowledge-Based Systems}, vol. 164, pp. 371--377, 2019.

\bibitem{joint1}
Y.~Pang, J.~Xie, F.~Nie, and X.~Li, ``Spectral clustering by joint spectral embedding and spectral rotation,'' \emph{IEEE Transactions on Cybernetics}, vol.~50, no.~1, pp. 247--258, 2020.

\bibitem{joint2}
Z.~Wang, Z.~Li, R.~Wang, F.~Nie, and X.~Li, ``Large graph clustering with simultaneous spectral embedding and discretization,'' \emph{IEEE Transactions on Pattern Analysis and Machine Intelligence}, vol.~43, no.~12, pp. 4426--4440, 2021.

\bibitem{ncut}
J.~Shi and J.~Malik, ``Normalized cuts and image segmentation,'' \emph{IEEE Transactions on Pattern Analysis and Machine Intelligence}, vol.~22, no.~8, pp. 888--905, 2000.

\bibitem{anchor1}
K.~Song, X.~Yao, F.~Nie, X.~Li, and M.~Xu, ``Weighted bilateral k-means algorithm for fast co-clustering and fast spectral clustering,'' \emph{Pattern Recognition}, vol. 109, p. 107560, 2021.

\bibitem{anchor2}
Y.~Zhao, Y.~Yuan, and Q.~Wang, ``Fast spectral clustering for unsupervised hyperspectral image classification,'' \emph{Remote Sensing}, vol.~11, no.~4, p. 399, 2019.

\bibitem{fcdmf}
F.~Nie, S.~Pei, R.~Wang, and X.~Li, ``Fast clustering with co-clustering via discrete non-negative matrix factorization for image identification,'' in \emph{ICASSP 2020 - 2020 IEEE International Conference on Acoustics, Speech and Signal Processing (ICASSP)}, 2020, pp. 2073--2077.

\bibitem{Ny1}
C.~Fowlkes, S.~Belongie, F.~Chung, and J.~Malik, ``Spectral grouping using the nystrom method,'' \emph{IEEE Transactions on Pattern Analysis and Machine Intelligence}, vol.~26, no.~2, pp. 214--225, 2004.

\bibitem{Nyfast1}
M.~Vladymyrov and M.~Carreira-Perpinan, ``The variational nystrom method for large-scale spectral problems,'' in \emph{Proceedings of The 33rd International Conference on Machine Learning}, vol.~48.\hskip 1em plus 0.5em minus 0.4em\relax PMLR, 2016, pp. 211--220.

\bibitem{Nyfast2}
H.~Jia, L.~Wang, H.~Song, Q.~Mao, and S.~Ding, ``An efficient nystr{\"o}m spectral clustering algorithm using incomplete cholesky decomposition,'' \emph{Expert Systems with Applications}, vol. 186, p. 115813, 2021.

\bibitem{FINCH}
M.~S. Sarfraz, V.~Sharma, and R.~Stiefelhagen, ``Efficient parameter-free clustering using first neighbor relations,'' in \emph{{IEEE} Conference on Computer Vision and Pattern Recognition, {CVPR} 2019, Long Beach, CA, USA, June 16-20, 2019}.\hskip 1em plus 0.5em minus 0.4em\relax Computer Vision Foundation / {IEEE}, 2019, pp. 8934--8943.

\bibitem{MCEMS}
T.~Li, A.~Rezaeipanah, and E.~M. T.~E. Din, ``An ensemble agglomerative hierarchical clustering algorithm based on clusters clustering technique and the novel similarity measurement,'' \emph{J. King Saud Univ. Comput. Inf. Sci.}, vol.~34, no. 6 Part {B}, pp. 3828--3842, 2022.

\bibitem{hc1}
V.~Cohen{-}Addad, V.~Kanade, F.~Mallmann{-}Trenn, and C.~Mathieu, ``Hierarchical clustering: Objective functions and algorithms,'' in \emph{Proceedings of the Twenty-Ninth Annual {ACM-SIAM} Symposium on Discrete Algorithms, {SODA}, 2018}, A.~Czumaj, Ed.\hskip 1em plus 0.5em minus 0.4em\relax {SIAM}, 2018, pp. 378--397.

\bibitem{hc3}
S.~Pasupathi, V.~Shanmuganathan, M.~Kaliappan, Y.~H. Robinson, and M.~Kim, ``Trend analysis using agglomerative hierarchical clustering approach for time series big data,'' \emph{J. Supercomput.}, vol.~77, no.~7, pp. 6505--6524, 2021.

\bibitem{hc4}
M.~Charikar and V.~Chatziafratis, ``Approximate hierarchical clustering via sparsest cut and spreading metrics,'' in \emph{Proceedings of the Twenty-Eighth Annual {ACM-SIAM} Symposium on Discrete Algorithms, {SODA} 2017, Barcelona, Spain, Hotel Porta Fira, January 16-19}, P.~N. Klein, Ed.\hskip 1em plus 0.5em minus 0.4em\relax {SIAM}, 2017, pp. 841--854.

\bibitem{hc5}
M.~Charikar, V.~Chatziafratis, and R.~Niazadeh, ``Hierarchical clustering better than average-linkage,'' in \emph{Proceedings of the Thirtieth Annual {ACM-SIAM} Symposium on Discrete Algorithms, {SODA} 2019, San Diego, California, USA, January 6-9, 2019}, T.~M. Chan, Ed.\hskip 1em plus 0.5em minus 0.4em\relax {SIAM}, 2019, pp. 2291--2304.

\bibitem{ksums}
S.~Pei, F.~Nie, R.~Wang, and X.~Li, ``Efficient clustering based on a unified view of $ k $-means and ratio-cut,'' \emph{Advances in Neural Information Processing Systems}, vol.~33, pp. 14\,855--14\,866, 2020.

\bibitem{uspec}
D.~Huang, C.-D. Wang, J.-S. Wu, J.-H. Lai, and C.-K. Kwoh, ``Ultra-scalable spectral clustering and ensemble clustering,'' \emph{IEEE Transactions on Knowledge and Data Engineering}, vol.~32, no.~6, pp. 1212--1226, 2020.

\bibitem{sesr}
Z.~Wang, X.~Dai, P.~Zhu, R.~Wang, X.~Li, and F.~Nie, ``Fast optimization of spectral embedding and improved spectral rotation,'' \emph{IEEE Transactions on Knowledge and Data Engineering}, pp. 1--1, 2021.

\bibitem{AGCI}
Y.~Zhao, Y.~Yuan, and Q.~Wang, ``Fast spectral clustering for unsupervised hyperspectral image classification,'' \emph{Remote Sensing}, vol.~11, no.~4, 2019.

\bibitem{GANC}
S.~S. Tabatabaei, M.~Coates, and M.~G. Rabbat, ``{GANC:} greedy agglomerative normalized cut for graph clustering,'' \emph{Pattern Recognition}, vol.~45, no.~2, pp. 831--843, 2012.

\bibitem{CLR}
F.~Nie, X.~Wang, M.~Jordan, and H.~Huang, ``The constrained laplacian rank algorithm for graph-based clustering,'' in \emph{Proceedings of the AAAI conference on artificial intelligence}, vol.~30, no.~1, 2016.

\bibitem{kmpp}
D.~Arthur and S.~Vassilvitskii, ``k-means++: The advantages of careful seeding,'' Stanford, Tech. Rep., 2006.

\end{thebibliography}

\end{document}